\documentclass[conference]{IEEEtran}
\IEEEoverridecommandlockouts
\usepackage{CJKutf8}
\usepackage{cite}
\usepackage{url}
\usepackage{amsmath,amssymb,amsfonts}
\usepackage{graphicx}
\usepackage{textcomp}
\usepackage{multicol}
\usepackage{multirow}
\usepackage{xcolor}
\usepackage{makecell}
\usepackage{changepage}
\usepackage{threeparttable}
\usepackage{hhline}
\usepackage{subfigure}
\usepackage{float}
\usepackage[ruled,vlined,linesnumbered]{algorithm2e}
\def\BibTeX{{\rm B\kern-.05em{\sc i\kern-.025em b}\kern-.08em
		T\kern-.1667em\lower.7ex\hbox{E}\kern-.125emX}}
\makeatletter
\newcommand{\linebreakand}{%
\end{@IEEEauthorhalign}
\hfill\mbox{}\par
\mbox{}\hfill\begin{@IEEEauthorhalign}
}
\makeatother

\begin{document}
\begin{CJK}{UTF8}{gbsn}

\title{RDEx-SOP: Exploitation-Biased Reconstructed Differential Evolution for Fixed-Budget Bound-Constrained Single-Objective Optimization}

\author{
	\IEEEauthorblockN{
		Sichen Tao\textsuperscript{1,2},
		Yifei Yang\textsuperscript{3},
		Ruihan Zhao\textsuperscript{4,5},
		Kaiyu Wang\textsuperscript{6,7},
		Sicheng Liu\textsuperscript{8},
		Shangce Gao\textsuperscript{1}
	}
	\IEEEauthorblockA{\textsuperscript{1}Department of Engineering, University of Toyama, Toyama-shi 930-8555, Japan}
	\IEEEauthorblockA{\textsuperscript{2}Cyberscience Center, Tohoku University, Sendai-shi 980-8578, Japan}
	\IEEEauthorblockA{\textsuperscript{3}Faculty of Science and Technology, Hirosaki University, Hirosaki-shi 036-8560, Japan}
	\IEEEauthorblockA{\textsuperscript{4}Sino-German College of Applied Sciences, Tongji University, Shanghai 200092, China}
	\IEEEauthorblockA{\textsuperscript{5}School of Mechanical Engineering, Tongji University, Shanghai 200092, China}
	\IEEEauthorblockA{\textsuperscript{6}Chongqing Institute of Microelectronics Industry Technology,\\University of Electronic Science and Technology of China, Chongqing 401332, China}
	\IEEEauthorblockA{\textsuperscript{7}Artificial Intelligence and Big Data College,\\Chongqing Polytechnic University of Electronic Technology, Chongqing 401331, China}
	\IEEEauthorblockA{\textsuperscript{8}Department of Information Engineering, Yantai Vocational College, Yantai 264670, China}
	\IEEEauthorblockA{
		\{sichen.tao@tohoku.ac.jp, taosc73@hotmail.com\}; yyf7236@hirosaki-u.ac.jp;\\
		\{ruihan\_zhao@tongji.edu.cn, ruihan.z@outlook.com\};\\
		\{wangky@uestc.edu.cn, greskowky1996@163.com\};\\
		\{20250024@ytvc.edu.cn, lsctoyama2020@gmail.com\}; gaosc@eng.u-toyama.ac.jp
	}
}

\maketitle

\begin{abstract}
Bound-constrained single-objective numerical optimisation remains a key benchmark for assessing the robustness and efficiency of evolutionary algorithms.
This report documents RDEx-SOP, an exploitation-biased success-history differential evolution variant used in the IEEE CEC 2025 numerical optimisation competition (C06 special session).
RDEx-SOP combines success-history parameter adaptation, an exploitation-biased hybrid branch, and lightweight local perturbations to balance fast convergence and final solution quality under a strict evaluation budget.
We evaluate RDEx-SOP on the official CEC 2025 SOP benchmark with the U-score framework (Speed and Accuracy categories).
Experimental results show that RDEx-SOP achieves strong overall performance and statistically competitive final outcomes across the 29 benchmark functions.
\end{abstract}


\section{Introduction}
Differential evolution (DE) remains one of the most competitive paradigms for continuous black-box optimisation because its search directions are generated directly from population differences, which gives strong landscape adaptivity at relatively low algorithmic complexity~\cite{storn1997differential}. In single-objective bound-constrained optimisation, the most successful DE variants have typically emerged not from completely new frameworks, but from a sustained sequence of refinements to mutation bias, parameter adaptation, and population scheduling.

This development line is now fairly clear. JADE introduced the current-to-$p$best mutation and archive mechanism that became the backbone of many later DE variants~\cite{zhang2009jade}. SHADE then replaced manually tuned parameters with success-history memories, and L-SHADE further improved the line by coupling success-history adaptation with linear population size reduction~\cite{tanabe2013success,tanabe2014improving}. Subsequent variants such as iL-SHADE and jSO refined parameter control and mutation pressure, while LSHADE-RSP and iLSHADE-RSP strengthened selective pressure and local perturbation, respectively~\cite{brest2016shade,brest2017single,stanovov2018lshade,choi2021improved}. This progression also supports the broader conclusion that success-history control is one of the most effective and best-benchmarked parameter-control directions in DE~\cite{tanabe2020reviewing}.

Two more recent observations are especially relevant to competition settings. First, ordered mutation can deliberately increase exploitation pressure without discarding differential information, as demonstrated by EB-LSHADE~\cite{mohamed2019novel}. Second, the success rate itself can serve as a compact learning signal for scaling-factor control, as shown in the recent LSRTDE line~\cite{stanovov2024successrate,stanovov2024lsrtde}. However, these ingredients are usually studied in isolation, and many comparisons still emphasise final objective values more than the joint speed-accuracy trade-off that matters in fixed-budget competitions.

The CEC 2025 SOP track makes that trade-off explicit because algorithms are ranked under the U-score framework, which rewards both target-attainment speed and final solution quality under a fixed evaluation budget~\cite{price2023trial}. RDEx-SOP is designed for this setting as a reconstructed DE that selectively recombines the most effective ideas from the post-SHADE literature rather than proposing a single isolated operator. Its main contribution is a compact competition-oriented formulation that couples success-history memories, linear population reduction, exploitation-biased hybrid mutation, adaptive operator allocation, and lightweight Cauchy local perturbation, together with an empirical study showing that this reconstruction is highly competitive on the official CEC 2025 benchmark.
The source code for RDEx-SOP is publicly available on Sichen Tao's GitHub page: \url{https://github.com/SichenTao}.

\section{Reconstructed Differential Evolution with Exploitation-biased Hybridization (RDEx-SOP)}

\subsection{Problem Formulation}
The bound-constrained single-objective optimisation problem considered in the SOP track can be written as:
\begin{equation}\label{eq:sop_problem}
\min_{x\in\mathbb{R}^D} f(x),\quad
\text{s.t. }\ell_j \le x_j \le u_j,\; j=1,2,\dots,D.
\end{equation}

\subsection{Overall Framework}
RDEx-SOP is a success-history adaptive differential evolution method with three main design elements:
\begin{itemize}
\item an elite \emph{front} population with linear size reduction~\cite{tanabe2014improving};
\item a two-branch mutation mechanism (\emph{standard} branch and \emph{exploitation-biased} (EB) branch) with an adaptive hybrid rate, following the ordered-mutation intuition of EB-LSHADE~\cite{mohamed2019novel};
\item a light local perturbation operator based on the Cauchy distribution~\cite{choi2021improved}.
\end{itemize}

Let $\mathcal{F}^{(g)}=\{x_i^{(g)}\}_{i=1}^{N^{(g)}}$ denote the front at generation $g$, where $N^{(g)}$ is the current front size.
For each target vector $x_i^{(g)}$, RDEx-SOP samples control parameters $(F_i^{(g)},CR_i^{(g)})$, generates a donor vector $v_i^{(g)}$ using one of the two mutation branches, and then applies binomial crossover to obtain a trial $u_i^{(g)}$.
If $f(u_i^{(g)}) \le f(x_i^{(g)})$, the trial is accepted and contributes to the success-history update.

\subsection{Standard Branch Mutation and Crossover}
\subsubsection{Mutation}
In the standard branch, RDEx-SOP uses a current-to-$p$best/1-like operator with an additional difference vector:
\begin{equation}\label{eq:sop_std_mut}
v_i^{(g)} = x_i^{(g)} + F_i^{(g)}\left(x_{pbest}^{(g)}-x_i^{(g)}\right) + F_i^{(g)}\left(x_{r_1}^{(g)}-x_{r_2}^{(g)}\right),
\end{equation}
where $x_{pbest}^{(g)}$ is selected from the top $p^{(g)}$ individuals of $\mathcal{F}^{(g)}$, and $x_{r_1}^{(g)}$, $x_{r_2}^{(g)}$ are randomly selected vectors (with indices different from each other and from $i$).

The selection window size $p^{(g)}$ is dynamically adjusted according to the empirical \emph{success rate}:
\begin{equation}\label{eq:sop_psize}
p^{(g)} = \max\left(2,\left\lfloor N^{(g)}\cdot \xi \cdot \exp\left(-k\cdot SR^{(g)}\right)\right\rfloor\right),
\end{equation}
where $SR^{(g)}\in[0,1]$ is the fraction of successful trials within the front in generation $g$, and $(\xi,k)$ are constants.

\subsubsection{Binomial Crossover and Bound Handling}
RDEx-SOP applies binomial crossover to combine $v_i^{(g)}$ and $x_i^{(g)}$:
\begin{equation}\label{eq:sop_xover}
u^{(g)}_{i,j}=
\begin{cases}
v^{(g)}_{i,j}, & \text{if }\mathrm{rand}(0,1)<CR_i^{(g)}\ \text{or } j=j_{\mathrm{rand}},\\
x^{(g)}_{i,j}, & \text{otherwise},
\end{cases}
\end{equation}
where $j_{\mathrm{rand}}\in\{1,\dots,D\}$ guarantees that at least one dimension is inherited from the donor.
If a component violates the bounds, RDEx-SOP repairs it by resampling uniformly within $[\ell_j,u_j]$.

\subsection{Exploitation-biased (EB) Branch}
The EB branch uses an ordered donor set to bias the search toward fitter regions while keeping a differential term for diversity.
Let $\{x_a^{(g)},x_b^{(g)},x_c^{(g)}\}$ be a donor set (sampled similarly to $\{x_{pbest}^{(g)},x_{r_1}^{(g)},x_{r_2}^{(g)}\}$), and let $\{x_{\mathrm{best}}^{(g)},x_{\mathrm{mid}}^{(g)},x_{\mathrm{worst}}^{(g)}\}$ denote these three donors ordered by fitness (best = lowest objective).
The EB mutation is:
\begin{equation}\label{eq:sop_eb_mut}
v_i^{(g)} = x_i^{(g)} + F_i^{(g)}\left(x_{\mathrm{best}}^{(g)}-x_i^{(g)}\right) + F_i^{(g)}\left(x_{\mathrm{mid}}^{(g)}-x_{\mathrm{worst}}^{(g)}\right).
\end{equation}

RDEx-SOP maintains an adaptive hybrid rate $\rho_{\mathrm{EB}}^{(g)}\in[0,1]$ that controls how often the EB branch is used.
After each generation, $\rho_{\mathrm{EB}}^{(g)}$ is updated according to the relative fitness improvements contributed by EB-generated trials versus standard-branch trials:
\begin{equation}\label{eq:sop_rho_update}
\rho_{\mathrm{EB}}^{(g+1)}=\frac{\sum_{i\in\mathcal{S}_{\mathrm{EB}}^{(g)}}\Delta f_i}{\sum_{i\in\mathcal{S}_{\mathrm{EB}}^{(g)}}\Delta f_i + \sum_{i\in\mathcal{S}_{\mathrm{std}}^{(g)}}\Delta f_i},
\end{equation}
where $\Delta f_i = f(x_i^{(g)})-f(u_i^{(g)})$ and $\mathcal{S}_{\mathrm{EB}}^{(g)}$ and $\mathcal{S}_{\mathrm{std}}^{(g)}$ are the sets of successful trials generated by the EB and standard branches, respectively.

\subsection{Success-history Parameter Sampling and Update}
RDEx-SOP uses two memories, $M_F$ and $M_{CR}$, each with $H$ entries, to store reference values of successful parameters.
In each generation, a memory index is selected and perturbed to generate $(F_i^{(g)},CR_i^{(g)})$.
In the standard branch, $F_i^{(g)}$ is sampled from a truncated Gaussian distribution, whose mean depends on the current success rate in the spirit of recent success-rate-driven DE adaptation~\cite{stanovov2024successrate}:
\begin{equation}\label{eq:sop_meanF}
\mu_F^{(g)} = 0.4 + 0.25\cdot\tanh\left(5\cdot SR^{(g)}\right),
\end{equation}
and $CR_i^{(g)}$ is sampled from a truncated Gaussian distribution centered at the memory value.
In the EB branch, $F_i^{(g)}$ is sampled from a Cauchy distribution centered at the memory value (with a fallback reference value when needed), and $CR_i^{(g)}$ is sampled similarly with early-stage lower bounds to encourage larger crossovers.

After the generation, successful parameters update the memory via a weighted Lehmer mean.
Let $\{F_k\}_{k=1}^{|\mathcal{S}^{(g)}|}$ and $\{CR_k\}_{k=1}^{|\mathcal{S}^{(g)}|}$ denote parameters of successful trials, and let the weights be $w_k=\Delta f_k/\sum_{t}\Delta f_t$.
Then a memory entry is updated as:
\begin{equation}\label{eq:sop_memF}
M_F \leftarrow \frac{\sum_k w_k F_k^2}{\sum_k w_k F_k},
\end{equation}
\begin{equation}\label{eq:sop_memCR}
M_{CR} \leftarrow \frac{1}{2}\left(M_{CR}+\frac{\sum_k w_k CR_k^2}{\sum_k w_k CR_k}\right).
\end{equation}

\subsection{Cauchy Local Perturbation}
To encourage local refinement (especially when crossover does not modify certain coordinates), RDEx-SOP applies a Cauchy perturbation to non-crossover dimensions with probability $p_r$:
\begin{equation}\label{eq:sop_cauchy}
u^{(g)}_{i,j} \leftarrow \mathrm{Cauchy}\left(x^{(g)}_{i,j},\sigma_{\mathrm{loc}}\right).
\end{equation}
This operator introduces occasional heavy-tailed jumps while keeping most coordinates unchanged.

\subsection{Linear Population Size Reduction}
Following the L-SHADE line~\cite{tanabe2014improving}, RDEx-SOP linearly reduces the front size from $N_0$ to a small minimum value $N_{\min}$ as the evaluation budget is consumed:
\begin{equation}\label{eq:sop_lpsr}
N^{(g+1)} = \left\lfloor N_0 + (N_{\min}-N_0)\cdot \frac{\mathrm{NFE}}{\mathrm{MaxFE}} \right\rfloor.
\end{equation}

\subsection{Pseudocode}
Algorithm~\ref{alg:rdex_sop} summarizes the overall RDEx-SOP procedure.
\begin{algorithm}[t]
\caption{RDEx-SOP framework.}
\label{alg:rdex_sop}
\KwIn{Population size $N_0$, memory size $H$, evaluation budget $\mathrm{MaxFE}$.}
\KwOut{Best solution found.}
Initialise $N^{(0)}\leftarrow N_0$ and sample an initial front population $\mathcal{F}^{(0)}$\;
Evaluate all individuals and initialise memories $M_F$ and $M_{CR}$\;
\While{$\mathrm{NFE}<\mathrm{MaxFE}$}{
  \ForEach{$x_i\in\mathcal{F}^{(g)}$}{
    Sample $(F_i,CR_i)$ from success-history memories\;
    Select EB branch with probability $\rho_{\mathrm{EB}}^{(g)}$\;
    Generate donor $v_i$ via Eq.~(\ref{eq:sop_std_mut}) or Eq.~(\ref{eq:sop_eb_mut})\;
    Generate trial $u_i$ by crossover Eq.~(\ref{eq:sop_xover}) and apply repairs/perturbation\;
    Evaluate $f(u_i)$ and perform greedy selection\;
    Store successful $(F_i,CR_i,\Delta f_i)$\;
  }
  Update $\rho_{\mathrm{EB}}$ by Eq.~(\ref{eq:sop_rho_update}) and update $M_F,M_{CR}$ by Eq.~(\ref{eq:sop_memF})--(\ref{eq:sop_memCR})\;
  Reduce front size using Eq.~(\ref{eq:sop_lpsr})\;
  $g \leftarrow g+1$\;
}
\end{algorithm}

\section{Experimental Results}
\subsection{Benchmark Functions}
The CEC 2025 SOP benchmark suite consists of 29 bound-constrained test functions with dimension $D=30$ and default variable bounds $[-100,100]^D$.
Following the official competition protocol, each function is evaluated with 25 independent runs.
The maximum number of function evaluations is set to $\mathrm{MaxFE}=10000\times D$, and the platform records the best-so-far objective values at $1000$ evenly spaced checkpoints (i.e., every $10\times D$ evaluations) for convergence-speed evaluation.

\subsection{Parameter Settings}
Unless otherwise stated, we use the default parameter configuration of the reference RDEx-SOP implementation:
\begin{itemize}
\item initial population/front size $N_0=600$ and minimum size $N_{\min}=4$;
\item success-history memory size $H=5$;
\item initial EB hybrid rate $\rho_{\mathrm{EB}}^{(0)}=0.7$;
\item local Cauchy perturbation probability $p_r=0.1$ and scale $\sigma_{\mathrm{loc}}=0.1$;
\item Gaussian perturbation scale for the standard-branch $F$: $\sigma_F=0.02$;
\item selection-pressure parameters $(\xi,k)=(0.7,7.0)$ in Eq.~(\ref{eq:sop_psize}).
\end{itemize}

\subsection{Experimental Settings}
RDEx-SOP is evaluated with the official U-score framework using the median target setting.
The U-score evaluation consists of two categories:
\textbf{Speed} (performance over checkpoints) and \textbf{Accuracy} (final-stage performance).
We compare RDEx-SOP with a selected set of representative competitors available in this repository: LSRTDE, jSOa, IEACOP, mLSHADE-LR, and RDE.

\subsection{Statistical Results}
	\subsubsection{Overall U-score Results}
Table~\ref{tab:cec2025_sop_scores} reports the overall U-score results for the selected algorithm set.
% --- SOP scores table (median target; 29 problems; 25 runs) ---
\begin{table}[t]
 \centering
 \caption{CEC 2025 SOP evaluation (median target): overall scores over 29 problems and 25 runs for the selected algorithm set.}
 \label{tab:cec2025_sop_scores}
 \scriptsize
 \setlength{\tabcolsep}{3pt}
 \begin{tabular}{|c|l|r|r|r|r|}
  \hline
  Rank & Algorithm & Total Score & Avg Score/Prob. & Speed & Accuracy \\ \hline
  1 & RDEx & 81229.5 & 2801.02 & 76490.5 & 4739.0 \\ \hline
  2 & LSRTDE & 73664.5 & 2540.16 & 68007.5 & 5657.0 \\ \hline
  3 & RDE & 69459.0 & 2395.14 & 58066.0 & 11393.0 \\ \hline
  4 & jSOa & 44653.0 & 1539.76 & 20984.0 & 23669.0 \\ \hline
  5 & mLSHADE\_LR & 44106.5 & 1520.91 & 25274.5 & 18832.0 \\ \hline
  6 & IEACOP & 15312.5 & 528.02 & 5635.5 & 9677.0 \\ \hline
 \end{tabular}
\end{table}

RDEx-SOP achieves the highest total score and the best overall rank mainly due to its leading performance in the Speed category, which is consistent with the design goal of accelerating convergence through the EB hybrid branch.

	\subsubsection{Per-function Statistics and Wilcoxon Tests}
	Due to the strict 4-page submission limit, detailed per-function statistics (means/SD and per-function Wilcoxon tests) are provided in the supplementary appendix of the full report.

	In terms of per-function Wilcoxon wins/ties/losses on final values, RDEx-SOP obtains 4/24/1 against LSRTDE, 21/5/3 against jSOa, and 27/1/1 against IEACOP, indicating that its final-stage accuracy is comparable to strong baselines while showing consistent improvements over several competitors.
	To make the pairwise comparisons more rigorous and informative, Table~\ref{tab:sop_acc_summary} further reports Holm-corrected W/T/L counts across the 29 functions (to control the family-wise error rate) and the Vargha--Delaney $A_{12}$ effect size for Final, TTT, and AUC (larger is better for minimization).
	\begin{table*}[t]
\centering
\caption{Pairwise summary over the 29 CEC2025 SOP functions (25 runs). For each metric (Final, TTT, and AUC), we report uncorrected per-function Wilcoxon W/T/L at $\alpha=0.05$, Holm-corrected W/T/L across the 29 functions, and the median Vargha--Delaney $A_{12}$ effect size (larger is better for minimization).}
\label{tab:sop_acc_summary}
\scriptsize
\setlength{\tabcolsep}{3pt}
\begin{tabular}{|l|ccc|ccc|ccc|}
\hline
 \multirow{2}{*}{Competitor} & \multicolumn{3}{c|}{Final} & \multicolumn{3}{c|}{TTT} & \multicolumn{3}{c|}{AUC} \\ \cline{2-10}
  & W/T/L & Holm & $A_{12}$ & W/T/L & Holm & $A_{12}$ & W/T/L & Holm & $A_{12}$ \\ \hline
 LSRTDE & 4/24/1 & 2/27/0 & 0.48 & 17/12/0 & 10/19/0 & 0.73 & 23/6/0 & 21/8/0 & 0.79 \\ \hline
 jSOa & 21/5/3 & 21/5/3 & 0.92 & 23/3/3 & 23/3/3 & 0.98 & 27/2/0 & 27/2/0 & 1.00 \\ \hline
 IEACOP & 27/1/1 & 27/1/1 & 1.00 & 25/3/1 & 25/3/1 & 1.00 & 23/0/6 & 23/0/6 & 1.00 \\ \hline
 mLSHADE-LR & 21/4/4 & 21/4/4 & 0.91 & 23/3/3 & 23/3/3 & 0.98 & 23/2/4 & 23/2/4 & 1.00 \\ \hline
 RDE & 17/6/6 & 17/6/6 & 0.83 & 18/2/9 & 18/2/9 & 0.93 & 19/2/8 & 19/2/8 & 1.00 \\ \hline
\end{tabular}
\end{table*}

	As complementary multi-algorithm comparisons, we report Friedman tests on per-function medians for final values, TTT, and AUC, together with the resulting average ranks (Table~\ref{tab:sop_friedman_summary}).
	\begin{table}[t]
\centering
\caption{Friedman tests on per-function medians over the 29 CEC2025 SOP functions (25 runs). Final: $\chi^2=77.67$, $df=5$, $p=2.57E-15$; TTT: $\chi^2=63.85$, $df=5$, $p=1.94E-12$; AUC: $\chi^2=60.07$, $df=5$, $p=1.17E-11$. Lower average rank indicates better performance for each metric.}
\label{tab:sop_friedman_summary}
\scriptsize
\setlength{\tabcolsep}{4pt}
\begin{tabular}{|l|c|c|c|}
\hline
 Algorithm & Final & TTT & AUC \\ \hline
 RDEx-SOP & \textbf{2.26} & \textbf{1.84} & \textbf{1.76} \\ \hline
 LSRTDE & 2.29 & 2.69 & 2.72 \\ \hline
 jSOa & 3.79 & 4.43 & 4.97 \\ \hline
 IEACOP & 5.83 & 4.95 & 4.03 \\ \hline
 mLSHADE-LR & 4.07 & 4.36 & 4.48 \\ \hline
 RDE & 2.76 & 2.72 & 3.03 \\ \hline
\end{tabular}
\end{table}

	To better reflect anytime convergence behaviour (which dominates the U-score Speed category), we additionally report two speed-focused metrics based on the same median target:
	(i) the time-to-target (TTT), defined as the first checkpoint index where a run reaches the target, and
	(ii) an AUC measure computed over 1000 checkpoints as $\log_{10}(1+\max(f_t-\mathrm{TGT},0))$.
Due to the strict 4-page submission limit, detailed per-function TTT/AUC tables are provided in the supplementary appendix of the full report.
In particular, RDEx-SOP shows consistent speed advantages over LSRTDE (TTT: 17/12/0; AUC: 23/6/0), which helps explain its superior U-score ranking even when final-value differences are often not statistically significant.

\subsection{Algorithm Complexity}
Let $D$ be the problem dimension, $N^{(g)}$ the current front size, and $\mathrm{MaxFE}$ the evaluation budget.
Let $T_f$ be the average cost of a single objective evaluation.
Per generation, RDEx-SOP performs $O(N^{(g)}D)$ arithmetic operations for mutation/crossover/repair and typically $O(N^{(g)}\log N^{(g)})$ operations for ranking and sampling from the front.
The dominant cost is the evaluation cost $O(\mathrm{MaxFE}\cdot T_f)$, while the algorithmic overhead is approximately $O(\mathrm{MaxFE}\cdot D)$ under the standard configuration with $N^{(g)}=O(D)$.
Therefore, for typical CEC benchmarks where evaluation is more expensive than vector operations, the overall runtime is mainly governed by $\mathrm{MaxFE}$ and the objective-function cost.

\section{Conclusion}
This report presented RDEx-SOP and its evaluation on the CEC 2025 SOP benchmark suite.
The results demonstrate strong overall U-score performance and competitive final-value statistics across the 29 problems, supporting the effectiveness of combining success-history adaptation with an exploitation-biased hybrid search mechanism.

\section*{Acknowledgment}
This research was partially supported by the Japan Society for the Promotion of Science (JSPS) KAKENHI under Grant JP22H03643, Japan Science and Technology Agency (JST) Support for Pioneering Research Initiated by the Next Generation (SPRING) under Grant JPMJSP2145, and JST through the Establishment of University Fellowships towards the Creation of Science Technology Innovation under Grant JPMJFS2115.

\bibliographystyle{IEEEtran}
\bibliography{references}

\onecolumn
% Appendix tables (supplementary large tables).

\appendices
\section{Supplementary Per-function Tables}
\label{app:sop_tables}

\subsection{Final-value statistics}
\begin{table}[H]
\centering
\caption{Experimental comparison results between RDEx-SOP and other competitors on the 29 benchmark functions in CEC2025 SOP.}
\label{tab:sop_exp_results}
\scriptsize
\renewcommand{\arraystretch}{0.9}
\setlength{\tabcolsep}{3pt}
\begin{tabular}{|c|cc|ccc|ccc|ccc|ccc|ccc|}
\hline
 \multirow{2}{*}{Problem} & \multicolumn{2}{c|}{RDEx-SOP} & \multicolumn{3}{c|}{LSRTDE} & \multicolumn{3}{c|}{jSOa} & \multicolumn{3}{c|}{IEACOP} & \multicolumn{3}{c|}{mLSHADE-LR} & \multicolumn{3}{c|}{RDE} \\ \cline{2-18}
  & Mean & SD & Mean & SD & W & Mean & SD & W & Mean & SD & W & Mean & SD & W & Mean & SD & W \\ \hline
 1 & \textbf{0.00E+00} & \textbf{0.00E+00} & \textbf{0.00E+00} & \textbf{0.00E+00} & = & 4.55E-15 & 6.63E-15 & = & 2.85E-06 & 8.17E-07 & + & \textbf{0.00E+00} & \textbf{0.00E+00} & = & \textbf{0.00E+00} & \textbf{0.00E+00} & = \\
 2 & \textbf{0.00E+00} & \textbf{0.00E+00} & \textbf{0.00E+00} & \textbf{0.00E+00} & = & 4.32E-14 & 2.43E-14 & + & 1.20E-06 & 1.68E-06 & + & 5.00E-14 & 2.45E-14 & + & \textbf{0.00E+00} & \textbf{0.00E+00} & = \\
 3 & 5.90E+01 & 1.51E+00 & 5.66E+01 & 1.08E+01 & = & 5.86E+01 & 7.11E-15 & - & \textbf{1.91E+00} & \textbf{1.99E+00} & - & 6.93E+00 & 1.49E+01 & - & 2.18E+01 & 1.22E+00 & - \\
 4 & 3.82E+00 & 1.90E+00 & \textbf{3.10E+00} & \textbf{1.39E+00} & = & 1.07E+01 & 2.17E+00 & + & 2.99E+01 & 7.31E+00 & + & 8.08E+00 & 3.07E+00 & + & 7.72E+00 & 1.57E+00 & + \\
 5 & 2.36E-07 & 5.47E-07 & 5.89E-08 & 1.70E-07 & = & 1.23E-13 & 3.08E-14 & = & 1.32E-01 & 2.05E-01 & + & 2.95E-03 & 1.03E-02 & = & \textbf{0.00E+00} & \textbf{0.00E+00} & = \\
 6 & \textbf{3.63E+01} & \textbf{2.31E+00} & 3.75E+01 & 9.42E+00 & = & 4.07E+01 & 2.12E+00 & + & 5.39E+01 & 5.75E+00 & + & 3.98E+01 & 3.11E+00 & + & 3.87E+01 & 2.19E+00 & + \\
 7 & 2.39E+00 & 1.43E+00 & \textbf{2.27E+00} & \textbf{1.75E+00} & = & 1.11E+01 & 2.00E+00 & + & 2.75E+01 & 6.33E+00 & + & 7.95E+00 & 2.61E+00 & + & 7.40E+00 & 1.87E+00 & + \\
 8 & \textbf{0.00E+00} & \textbf{0.00E+00} & \textbf{0.00E+00} & \textbf{0.00E+00} & = & \textbf{0.00E+00} & \textbf{0.00E+00} & = & 2.38E+00 & 9.73E+00 & + & \textbf{0.00E+00} & \textbf{0.00E+00} & = & \textbf{0.00E+00} & \textbf{0.00E+00} & = \\
 9 & 4.61E+02 & 3.56E+02 & \textbf{3.25E+02} & \textbf{2.20E+02} & = & 1.40E+03 & 2.37E+02 & + & 2.64E+03 & 4.67E+02 & + & 1.47E+03 & 2.86E+02 & + & 1.41E+03 & 2.29E+02 & + \\
 10 & 1.63E+00 & 1.76E+00 & \textbf{1.47E+00} & \textbf{1.87E+00} & = & 7.38E+00 & 1.62E+01 & + & 3.16E+01 & 1.34E+01 & + & 8.12E+00 & 1.08E+01 & + & 3.20E+00 & 2.80E+00 & = \\
 11 & 2.14E+00 & 2.06E+00 & \textbf{1.55E+00} & \textbf{1.71E+00} & = & 1.74E+02 & 9.50E+01 & + & 1.50E+03 & 4.76E+02 & + & 1.18E+03 & 4.25E+02 & + & 2.34E+02 & 1.20E+02 & + \\
 12 & \textbf{9.43E+00} & \textbf{6.60E+00} & 1.15E+01 & 5.59E+00 & = & 1.95E+01 & 2.45E+00 & + & 1.58E+02 & 1.91E+02 & + & 1.97E+01 & 8.35E+00 & + & 1.52E+01 & 5.03E+00 & + \\
 13 & 8.17E+00 & 9.84E+00 & \textbf{1.01E+00} & \textbf{3.95E+00} & - & 1.24E+01 & 9.26E+00 & + & 1.02E+02 & 2.75E+01 & + & 2.28E+01 & 3.35E+00 & + & 1.89E+01 & 7.20E+00 & + \\
 14 & 4.88E-01 & 2.67E-01 & \textbf{4.45E-01} & \textbf{1.67E-01} & = & 3.27E+00 & 1.64E+00 & + & 7.86E+01 & 8.31E+01 & + & 1.25E+01 & 1.29E+01 & + & 1.36E+00 & 8.64E-01 & + \\
 15 & 5.58E+00 & 4.64E+00 & \textbf{4.57E+00} & \textbf{3.37E+00} & = & 7.83E+01 & 8.52E+01 & + & 3.75E+02 & 1.49E+02 & + & 5.57E+01 & 5.71E+01 & + & 2.69E+01 & 4.14E+01 & + \\
 16 & \textbf{1.56E+01} & \textbf{1.09E+01} & 2.11E+01 & 6.36E+00 & = & 3.36E+01 & 8.12E+00 & + & 1.08E+02 & 6.70E+01 & + & 3.60E+01 & 8.99E+00 & + & 3.01E+01 & 1.07E+01 & + \\
 17 & \textbf{1.17E+01} & \textbf{9.95E+00} & 1.34E+01 & 9.67E+00 & + & 1.84E+01 & 6.27E+00 & + & 2.28E+02 & 5.69E+01 & + & 3.05E+01 & 7.41E+00 & + & 2.08E+01 & 4.43E-01 & + \\
 18 & \textbf{2.02E+00} & \textbf{6.12E-01} & 2.09E+00 & 6.08E-01 & = & 4.81E+00 & 1.35E+00 & + & 7.72E+01 & 7.24E+01 & + & 1.09E+01 & 4.78E+00 & + & 3.11E+00 & 9.89E-01 & + \\
 19 & \textbf{1.32E+01} & \textbf{3.53E+01} & 1.59E+01 & 3.50E+01 & = & 2.99E+01 & 6.75E+00 & + & 1.84E+02 & 3.12E+01 & + & 4.15E+01 & 1.02E+01 & + & 2.54E+01 & 5.17E+00 & + \\
 20 & \textbf{2.02E+02} & \textbf{1.60E+00} & 2.03E+02 & 1.75E+00 & = & 2.11E+02 & 2.06E+00 & + & 2.24E+02 & 6.85E+00 & + & 2.08E+02 & 1.98E+00 & + & 2.07E+02 & 1.93E+00 & + \\
 21 & 1.00E+02 & 0.00E+00 & 1.00E+02 & 1.67E-13 & + & 1.00E+02 & 3.23E-13 & - & 1.00E+02 & 1.56E-08 & + & \textbf{1.00E+02} & \textbf{0.00E+00} & - & \textbf{1.00E+02} & \textbf{0.00E+00} & - \\
 22 & \textbf{3.40E+02} & \textbf{2.70E+00} & 3.40E+02 & 3.70E+00 & = & 3.54E+02 & 3.97E+00 & + & 3.77E+02 & 1.16E+01 & + & 3.57E+02 & 7.35E+00 & + & 3.47E+02 & 2.96E+00 & + \\
 23 & 4.17E+02 & 3.07E+00 & \textbf{4.15E+02} & \textbf{3.93E+00} & = & 4.28E+02 & 2.60E+00 & + & 4.42E+02 & 3.89E+00 & + & 4.25E+02 & 4.35E+00 & + & 4.23E+02 & 1.68E+00 & + \\
 24 & 3.87E+02 & 7.69E-03 & 3.87E+02 & 3.66E-03 & = & 3.87E+02 & 6.67E-03 & = & 3.86E+02 & 1.65E+00 & = & 3.81E+02 & 2.62E+00 & - & \textbf{3.79E+02} & \textbf{1.67E-01} & - \\
 25 & \textbf{5.89E+02} & \textbf{1.32E+02} & 6.79E+02 & 1.14E+02 & + & 9.43E+02 & 3.54E+01 & + & 1.06E+03 & 4.73E+02 & + & 9.91E+02 & 7.41E+01 & + & 8.99E+02 & 3.02E+01 & + \\
 26 & \textbf{4.70E+02} & \textbf{3.35E+00} & 4.71E+02 & 3.57E+00 & = & 5.00E+02 & 5.00E+00 & + & 5.12E+02 & 6.77E+00 & + & 5.04E+02 & 8.88E+00 & + & 4.71E+02 & 5.93E+00 & = \\
 27 & 3.00E+02 & 2.27E-13 & 3.00E+02 & 1.83E-13 & + & 3.09E+02 & 3.09E+01 & = & 3.19E+02 & 4.34E+01 & + & \textbf{3.00E+02} & \textbf{0.00E+00} & - & 3.09E+02 & 3.09E+01 & - \\
 28 & 4.02E+02 & 6.74E+00 & 4.01E+02 & 6.37E+00 & = & 4.31E+02 & 8.68E+00 & + & 5.19E+02 & 4.86E+01 & + & 4.27E+02 & 2.02E+01 & + & \textbf{3.66E+02} & \textbf{4.64E+01} & - \\
 29 & 1.98E+03 & 1.07E+01 & 1.98E+03 & 1.35E+01 & = & 1.97E+03 & 1.50E+01 & - & 3.22E+03 & 5.44E+02 & + & 1.95E+03 & 1.65E+02 & = & \textbf{7.66E+02} & \textbf{3.60E+02} & - \\
\hline
 W/T/L & \multicolumn{2}{c|}{$-/-/-$} & \multicolumn{3}{c|}{4/24/1} & \multicolumn{3}{c|}{21/5/3} & \multicolumn{3}{c|}{27/1/1} & \multicolumn{3}{c|}{21/4/4} & \multicolumn{3}{c|}{17/6/6} \\
\hline
\end{tabular}
\end{table}

\subsection{Speed-side per-function statistics}
\begin{table}[H]
\centering
\caption{Time-to-target comparison on the 29 CEC2025 SOP functions. The time-to-target is the first checkpoint index (1--1000) where the run reaches the median target (smaller is better); runs that never reach the target are assigned 1001.}
\label{tab:sop_ttt_results}
\scriptsize
\renewcommand{\arraystretch}{0.9}
\setlength{\tabcolsep}{3pt}
\begin{tabular}{|c|cc|ccc|ccc|ccc|ccc|ccc|}
\hline
 \multirow{2}{*}{Problem} & \multicolumn{2}{c|}{RDEx-SOP} & \multicolumn{3}{c|}{LSRTDE} & \multicolumn{3}{c|}{jSOa} & \multicolumn{3}{c|}{IEACOP} & \multicolumn{3}{c|}{mLSHADE-LR} & \multicolumn{3}{c|}{RDE} \\ \cline{2-18}
  & Mean & SD & Mean & SD & W & Mean & SD & W & Mean & SD & W & Mean & SD & W & Mean & SD & W \\ \hline
 1 & 538.2 & 4.1 & 578.2 & 3.8 & + & 785.2 & 148.1 & + & 1001.0 & 0.0 & + & 805.5 & 70.6 & + & \textbf{477.0} & \textbf{5.6} & - \\
 2 & 485.9 & 5.5 & 514.0 & 5.0 & + & 920.7 & 143.0 & + & 1001.0 & 0.0 & + & 980.5 & 51.5 & + & \textbf{412.4} & \textbf{4.5} & - \\
 3 & 1001.0 & 0.0 & 975.8 & 123.7 & = & 1001.0 & 0.0 & = & \textbf{114.0} & \textbf{41.1} & - & 811.2 & 93.9 & - & 739.9 & 42.7 & - \\
 4 & \textbf{675.7} & \textbf{73.9} & 705.8 & 38.3 & + & 999.0 & 7.7 & + & 1001.0 & 0.0 & + & 971.5 & 38.5 & + & 981.8 & 21.2 & + \\
 5 & 846.8 & 136.9 & 814.3 & 116.5 & = & 811.7 & 65.2 & = & 1001.0 & 0.0 & + & 1001.0 & 0.0 & + & \textbf{607.2} & \textbf{4.6} & - \\
 6 & \textbf{760.0} & \textbf{105.7} & 874.6 & 58.7 & + & 995.4 & 10.8 & + & 1001.0 & 0.0 & + & 991.4 & 18.7 & + & 981.2 & 19.2 & + \\
 7 & \textbf{649.8} & \textbf{38.6} & 697.0 & 37.2 & + & 998.8 & 6.2 & + & 1001.0 & 0.0 & + & 975.0 & 48.3 & + & 985.3 & 16.5 & + \\
 8 & 454.6 & 5.0 & 485.1 & 3.2 & + & 567.8 & 3.1 & + & 1001.0 & 0.0 & + & 431.0 & 142.2 & = & \textbf{146.8} & \textbf{3.2} & - \\
 9 & \textbf{701.2} & \textbf{33.5} & 728.8 & 46.2 & = & 992.7 & 14.7 & + & 998.2 & 13.9 & + & 970.0 & 51.6 & + & 988.2 & 19.3 & + \\
 10 & 611.5 & 253.7 & \textbf{556.7} & \textbf{142.8} & = & 949.2 & 47.8 & + & 1001.0 & 0.0 & + & 971.1 & 53.8 & + & 824.8 & 132.1 & + \\
 11 & \textbf{553.4} & \textbf{49.6} & 568.0 & 35.4 & = & 769.2 & 193.1 & + & 1001.0 & 0.0 & + & 1001.0 & 0.0 & + & 771.3 & 283.0 & = \\
 12 & \textbf{595.2} & \textbf{82.8} & 674.2 & 84.9 & + & 997.4 & 8.7 & + & 1001.0 & 0.0 & + & 964.4 & 68.7 & + & 906.4 & 95.8 & + \\
 13 & 537.6 & 74.5 & \textbf{530.9} & \textbf{38.4} & = & 943.5 & 52.2 & + & 1001.0 & 0.0 & + & 994.4 & 25.9 & + & 944.4 & 96.3 & + \\
 14 & 540.7 & 52.1 & \textbf{535.0} & \textbf{30.8} & = & 984.8 & 38.5 & + & 1001.0 & 0.0 & + & 995.2 & 28.4 & + & 875.6 & 92.2 & + \\
 15 & 636.8 & 148.9 & \textbf{618.1} & \textbf{70.4} & = & 1001.0 & 0.0 & + & 1001.0 & 0.0 & + & 964.9 & 53.3 & + & 942.0 & 67.3 & + \\
 16 & \textbf{706.4} & \textbf{84.9} & 712.0 & 42.7 & = & 996.3 & 9.1 & + & 1001.0 & 0.0 & + & 986.8 & 31.2 & + & 984.1 & 22.9 & + \\
 17 & \textbf{615.2} & \textbf{126.6} & 781.2 & 239.4 & + & 938.6 & 76.9 & + & 1001.0 & 0.0 & + & 1001.0 & 0.0 & + & 911.3 & 58.8 & + \\
 18 & \textbf{620.9} & \textbf{30.1} & 645.4 & 39.5 & + & 997.4 & 6.7 & + & 1001.0 & 0.0 & + & 1001.0 & 0.0 & + & 903.7 & 64.9 & + \\
 19 & \textbf{576.2} & \textbf{129.0} & 612.4 & 119.5 & + & 994.5 & 11.5 & + & 1001.0 & 0.0 & + & 1000.5 & 1.8 & + & 952.6 & 44.0 & + \\
 20 & \textbf{547.3} & \textbf{29.8} & 595.7 & 32.5 & + & 1001.0 & 0.0 & + & 1001.0 & 0.0 & + & 979.9 & 35.6 & + & 973.8 & 27.6 & + \\
 21 & 689.1 & 9.8 & 984.8 & 38.2 & + & 704.9 & 109.4 & - & 1001.0 & 0.0 & + & \textbf{321.1} & \textbf{71.8} & - & 325.3 & 2.6 & - \\
 22 & \textbf{422.2} & \textbf{21.2} & 444.4 & 27.5 & + & 994.5 & 22.0 & + & 1001.0 & 0.0 & + & 945.9 & 122.9 & + & 952.6 & 36.4 & + \\
 23 & \textbf{400.5} & \textbf{24.1} & 425.5 & 25.1 & + & 998.5 & 12.1 & + & 1001.0 & 0.0 & + & 810.4 & 257.0 & + & 955.7 & 65.2 & + \\
 24 & 832.8 & 240.3 & 805.4 & 234.5 & = & 955.4 & 154.8 & = & 732.6 & 384.6 & = & 778.6 & 92.5 & = & \textbf{132.5} & \textbf{15.8} & - \\
 25 & \textbf{334.8} & \textbf{55.6} & 362.7 & 19.0 & + & 994.9 & 21.0 & + & 775.7 & 401.0 & + & 927.6 & 200.1 & + & 787.2 & 212.8 & + \\
 26 & \textbf{293.5} & \textbf{12.8} & 308.0 & 20.7 & + & 1001.0 & 0.0 & + & 1001.0 & 0.0 & + & 1001.0 & 0.0 & + & 392.9 & 198.5 & = \\
 27 & 906.0 & 114.3 & 1001.0 & 0.0 & = & 789.3 & 132.9 & - & 1001.0 & 0.0 & = & 769.3 & 60.8 & - & \textbf{500.7} & \textbf{152.2} & - \\
 28 & \textbf{631.4} & \textbf{82.4} & 671.6 & 104.3 & + & 1001.0 & 0.0 & + & 1001.0 & 0.0 & + & 983.3 & 39.7 & + & 909.6 & 58.0 & + \\
 29 & 905.8 & 200.9 & 817.5 & 249.1 & = & 742.2 & 245.0 & - & 1001.0 & 0.0 & = & 865.8 & 135.7 & = & \textbf{300.2} & \textbf{59.8} & - \\
\hline
 W/T/L & \multicolumn{2}{c|}{$-/-/-$} & \multicolumn{3}{c|}{17/12/0} & \multicolumn{3}{c|}{23/3/3} & \multicolumn{3}{c|}{25/3/1} & \multicolumn{3}{c|}{23/3/3} & \multicolumn{3}{c|}{18/2/9} \\
\hline
\end{tabular}
\end{table}

\begin{table}[H]
\centering
\caption{Anytime convergence comparison using AUC over 1000 checkpoints on the 29 CEC2025 SOP functions. For each run, AUC is computed as the mean of $\log_{10}(1+\max(f_t-\mathrm{TGT},0))$ across checkpoints (smaller is better).}
\label{tab:sop_auc_results}
\scriptsize
\renewcommand{\arraystretch}{0.9}
\setlength{\tabcolsep}{3pt}
\begin{tabular}{|c|cc|ccc|ccc|ccc|ccc|ccc|}
\hline
 \multirow{2}{*}{Problem} & \multicolumn{2}{c|}{RDEx-SOP} & \multicolumn{3}{c|}{LSRTDE} & \multicolumn{3}{c|}{jSOa} & \multicolumn{3}{c|}{IEACOP} & \multicolumn{3}{c|}{mLSHADE-LR} & \multicolumn{3}{c|}{RDE} \\ \cline{2-18}
  & Mean & SD & Mean & SD & W & Mean & SD & W & Mean & SD & W & Mean & SD & W & Mean & SD & W \\ \hline
 1 & 2.09 & 0.05 & 2.34 & 0.03 & + & 2.54 & 0.02 & + & \textbf{0.51} & \textbf{0.02} & - & 1.98 & 0.12 & - & 1.75 & 0.03 & - \\
 2 & 0.78 & 0.01 & 0.82 & 0.02 & + & 1.17 & 0.02 & + & 1.47 & 0.29 & + & \textbf{0.71} & \textbf{0.17} & = & 0.73 & 0.02 & - \\
 3 & 1.36 & 0.04 & 1.32 & 0.13 & = & 1.36 & 0.01 & = & \textbf{0.24} & \textbf{0.08} & - & 1.35 & 0.16 & = & 1.07 & 0.06 & - \\
 4 & \textbf{1.45} & \textbf{0.09} & 1.53 & 0.08 & + & 1.87 & 0.03 & + & 1.53 & 0.12 & + & 1.98 & 0.04 & + & 1.83 & 0.03 & + \\
 5 & 0.29 & 0.01 & 0.30 & 0.01 & + & 0.33 & 0.00 & + & 0.35 & 0.12 & + & \textbf{0.14} & \textbf{0.01} & - & 0.19 & 0.00 & - \\
 6 & 1.59 & 0.05 & 1.92 & 0.12 & + & 1.94 & 0.02 & + & \textbf{1.46} & \textbf{0.12} & - & 2.03 & 0.04 & + & 1.88 & 0.02 & + \\
 7 & \textbf{1.41} & \textbf{0.08} & 1.51 & 0.08 & + & 1.88 & 0.03 & + & 1.49 & 0.09 & + & 1.96 & 0.05 & + & 1.83 & 0.03 & + \\
 8 & 0.52 & 0.01 & 0.56 & 0.02 & + & 0.71 & 0.01 & + & 1.16 & 0.54 & + & \textbf{0.29} & \textbf{0.01} & - & 0.35 & 0.01 & - \\
 9 & \textbf{2.57} & \textbf{0.13} & 2.67 & 0.17 & = & 3.47 & 0.05 & + & 3.23 & 0.10 & + & 3.36 & 0.14 & + & 3.48 & 0.06 & + \\
 10 & \textbf{0.84} & \textbf{0.06} & 0.89 & 0.04 & + & 1.57 & 0.23 & + & 1.57 & 0.17 & + & 1.60 & 0.21 & + & 1.23 & 0.20 & + \\
 11 & \textbf{2.29} & \textbf{0.12} & 2.38 & 0.08 & + & 3.20 & 0.41 & + & 3.87 & 0.16 & + & 4.24 & 0.12 & + & 2.63 & 0.65 & = \\
 12 & \textbf{1.54} & \textbf{0.11} & 1.71 & 0.12 & + & 2.76 & 0.05 & + & 2.74 & 0.38 & + & 2.57 & 0.19 & + & 2.05 & 0.13 & + \\
 13 & \textbf{0.97} & \textbf{0.05} & 1.00 & 0.06 & = & 1.75 & 0.04 & + & 2.55 & 0.30 & + & 1.75 & 0.04 & + & 1.42 & 0.07 & + \\
 14 & \textbf{1.08} & \textbf{0.04} & 1.13 & 0.03 & + & 2.15 & 0.07 & + & 2.33 & 0.35 & + & 2.14 & 0.16 & + & 1.55 & 0.09 & + \\
 15 & \textbf{1.69} & \textbf{0.19} & 1.70 & 0.11 & = & 2.78 & 0.10 & + & 2.68 & 0.17 & + & 2.71 & 0.17 & + & 2.48 & 0.17 & + \\
 16 & \textbf{1.44} & \textbf{0.12} & 1.50 & 0.10 & = & 2.21 & 0.05 & + & 1.93 & 0.32 & + & 2.19 & 0.08 & + & 2.12 & 0.06 & + \\
 17 & \textbf{0.94} & \textbf{0.02} & 0.99 & 0.03 & + & 1.88 & 0.06 & + & 3.43 & 0.32 & + & 2.02 & 0.15 & + & 1.34 & 0.07 & + \\
 18 & \textbf{1.07} & \textbf{0.02} & 1.13 & 0.03 & + & 2.16 & 0.03 & + & 2.13 & 0.23 & + & 2.08 & 0.08 & + & 1.59 & 0.05 & + \\
 19 & \textbf{1.29} & \textbf{0.32} & 1.37 & 0.31 & + & 2.32 & 0.05 & + & 2.34 & 0.09 & + & 2.29 & 0.08 & + & 2.15 & 0.07 & + \\
 20 & \textbf{1.17} & \textbf{0.07} & 1.27 & 0.07 & + & 1.86 & 0.03 & + & 1.42 & 0.12 & + & 1.94 & 0.04 & + & 1.78 & 0.04 & + \\
 21 & 0.42 & 0.01 & 0.48 & 0.01 & + & 0.55 & 0.01 & + & \textbf{0.14} & \textbf{0.01} & - & 0.30 & 0.01 & - & 0.35 & 0.01 & - \\
 22 & \textbf{0.91} & \textbf{0.04} & 0.96 & 0.05 & + & 1.89 & 0.04 & + & 1.65 & 0.12 & + & 1.78 & 0.25 & + & 1.78 & 0.06 & + \\
 23 & \textbf{0.86} & \textbf{0.05} & 0.91 & 0.05 & + & 1.84 & 0.04 & + & 1.47 & 0.08 & + & 1.22 & 0.28 & + & 1.73 & 0.09 & + \\
 24 & 0.32 & 0.01 & 0.35 & 0.01 & + & 0.36 & 0.01 & + & 0.24 & 0.08 & - & 0.60 & 0.08 & + & \textbf{0.23} & \textbf{0.01} & - \\
 25 & \textbf{1.07} & \textbf{0.18} & 1.15 & 0.05 & + & 2.84 & 0.07 & + & 2.17 & 1.12 & + & 2.35 & 0.47 & + & 2.09 & 0.38 & + \\
 26 & 0.53 & 0.02 & 0.55 & 0.03 & + & 1.38 & 0.11 & + & 1.49 & 0.10 & + & 1.55 & 0.09 & + & \textbf{0.51} & \textbf{0.20} & = \\
 27 & \textbf{0.52} & \textbf{0.06} & 0.54 & 0.03 & + & 0.94 & 0.37 & + & 0.58 & 0.67 & - & 1.48 & 0.19 & + & 0.80 & 0.40 & + \\
 28 & \textbf{1.45} & \textbf{0.13} & 1.53 & 0.12 & + & 2.39 & 0.04 & + & 2.25 & 0.12 & + & 2.39 & 0.09 & + & 2.18 & 0.10 & + \\
 29 & 2.01 & 0.29 & 1.96 & 0.35 & = & 2.10 & 0.25 & = & 3.51 & 0.19 & + & 3.01 & 0.34 & + & \textbf{1.22} & \textbf{0.10} & - \\
\hline
 W/T/L & \multicolumn{2}{c|}{$-/-/-$} & \multicolumn{3}{c|}{23/6/0} & \multicolumn{3}{c|}{27/2/0} & \multicolumn{3}{c|}{23/0/6} & \multicolumn{3}{c|}{23/2/4} & \multicolumn{3}{c|}{19/2/8} \\
\hline
\end{tabular}
\end{table}

\end{CJK}
\end{document}